\documentclass{article}
\usepackage{spconf,amsmath,graphicx}
\usepackage{marvosym}
\usepackage[linesnumbered,ruled,vlined]{algorithm2e}
\usepackage{algpseudocode}
\usepackage[most]{tcolorbox}
\usepackage{algpseudocode}
\usepackage{bbold}
\usepackage{url}
\usepackage{mflogo}
\algnewcommand{\uIf}[2]{\State \algorithmicif\ #1\ \algorithmicthen\ #2}
\usepackage{enumitem}
\setlist{nosep, leftmargin=14pt}
\usepackage{mwe}

\usepackage{hyperref}
\title{GS-EMA: Integrating Gradient Surgery Exponential Moving Average with Boundary-Aware Contrastive Learning for Enhanced Domain Generalization in Aneurysm Segmentation}
\name{
    \begin{tabular}{c}
    Fengming Lin\textsuperscript{1 *},
    Yan Xia\textsuperscript{1 *}\thanks{\textasteriskcentered \quad Contribute equally to this work},
    Michael MacRaild\textsuperscript{1},
    Yash Deo\textsuperscript{1},
    Haoran Dou\textsuperscript{1},
    Qiongyao Liu\textsuperscript{1},
    Nina Cheng\textsuperscript{1}, \\
    Nishant Ravikumar\textsuperscript{1 \dag},
    Alejandro F. Frangi\textsuperscript{1 2 \dag}\thanks{\dag \quad Joint last authors} \thanks{\dag \quad AFF is supported by the Royal Academy of Engineering INSILEX Chair (CiET1919/19), UKRI Frontier Research Guarantee INSILICO (EP/Y030494/1), and EC Sixth Framework Programme @neurIST (FP6-2004-IST-4-027703).}
    \end{tabular}%
}
\address{
    \begin{tabular}{c}
    \textsuperscript{1} University of Leeds, \textsuperscript{2} University of Manchester  \quad
    \href{https://github.com/fmlinks/domain}{https://github.com/fmlinks/domain}
    \end{tabular}%
}

\begin{document}

\maketitle

\begin{abstract}
The automated segmentation of cerebral aneurysms is pivotal for accurate diagnosis and treatment planning. Confronted with significant domain shifts and class imbalance in 3D Rotational Angiography (3DRA) data from various medical institutions, the task becomes challenging. These shifts include differences in image appearance, intensity distribution, resolution, and aneurysm size, all of which complicate the segmentation process. To tackle these issues, we propose a novel domain generalization strategy that employs gradient surgery exponential moving average (GS-EMA) optimization technique coupled with boundary-aware contrastive learning (BACL). Our approach is distinct in its ability to adapt to new, unseen domains by learning domain-invariant features, thereby improving the robustness and accuracy of aneurysm segmentation across diverse clinical datasets. The results demonstrate that our proposed approach can extract more domain-invariant features, minimizing over-segmentation and capturing more complete aneurysm structures.

\end{abstract}

\keywords{Domain Generalization, Gradient Surgery, Contrastive Learning}

\section{Introduction}

\begin{figure}[htb]
\centerline{\includegraphics[width=\linewidth]{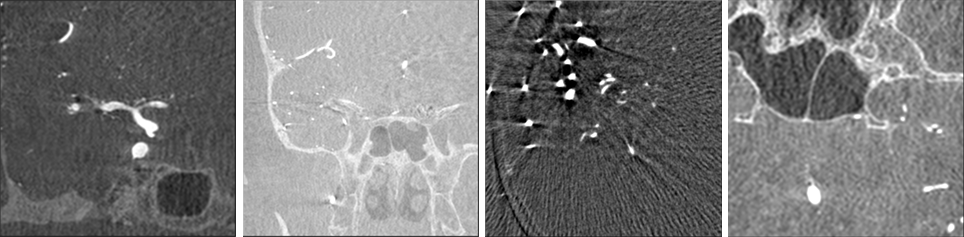}}
\caption{Illustration of the variability in imaging data quality from different medical centers.}
\label{fig1}
\end{figure}

The accurate segmentation of cerebral aneurysms is vital for diagnosing and treating patients effectively. This process is not just about detecting aneurysms early; it involves precise measurements of their size and shape, which are critical for formulating treatment plans \cite{liu2023hemodynamics, pappalardo2019silico}. However, the variability in imaging data quality (see Fig.\ref{fig1}) from different medical centers presents a significant challenge, complicating the segmentation process.

This variability necessitates a domain generalization (DG) approach, where a model trained on data from multiple sources can adapt to new, unseen domains. The diversity of multi-source data makes DG a daunting challenge in medical imaging, pushing the need for models that generalize well across different medical centers and data types.
% \vspace{-0.1cm}

Unlike traditional DG approaches such as domain alignment \cite{lu2022domain}, data augmentation \cite{liu2021feddg}, ensemble learning \cite{wang2020dofe}, self-supervised learning \cite{krishnan2022self}, disentangled representation learning \cite{zhang2022towards}, and others, our method takes a different approach. We enhance domain generalization by leveraging gradient
surgery exponential moving average (GS-EMA), offering an innovative solution to address DG challenges.

In deep learning, EMA is a frequently used technique for parameter averaging in models, aimed at enhancing the generalization performance and stability of the model.
In a teacher-student \cite{lin2023adaptive} network setup, the teacher network undergoes a process of parameter smoothing, driven by the student network. However, initially, there are no specific conditions set for this transfer. Consequently, all parameters learned by the student network, whether they are domain-invariant or domain-specific, are updated into the teacher network at some rate. This approach poses a challenge as it fails to distinguish between domain-invariant and domain-specific parameters. To address this issue, we introduce the concept of gradient surgery.

Deep neural networks are trained using gradient descent, where gradients guide the optimization process across the landscape defined by the loss function and training data. The gradient surgery framework \cite{yu2020gradient, mansilla2021domain} aims to resolve conflicts arising in multi-task learning. The conflicting gradients are typically averaged to obtain a final gradient for parameter updates. GSMorph \cite{dou2023gsmorph} propose alternative methods like normal vector projection to derive the ultimate gradient for parameter updates.
Instead of devising a new projection method as suggested by others, we approach the problem by analyzing the relationships between gradients to determine whether EMA parameter updates should occur.

Additionally, there is a class imbalance problem in 3D data segmentation due to the small proportion of aneurysms. After multiple downsampling steps, these small features are more likely to be overlooked in the latent space. To tackle this, we introduce the concept of boundary-awareness to traditional contrastive learning \cite{yang2022source}.

\textbf{Contributions}: Our study introduces innovative techniques that enhance model adaptability. We integrate gradient surgery with EMA updates, strengthening the ability of model to learn domain-invariant features. This novel approach promises to elevate the performance of DG tasks in medical imaging, ensuring that our model can generalize effectively to new datasets and medical centers. Additionally, we pioneer the use of boundary-aware contrastive learning, enabling our model to discern small target features especially for cerebral aneurysms. 
\vspace{-0.3cm}

\vspace{-0.1cm}
\section{Methods}
\begin{figure}[!ht]
\centerline{\includegraphics[width=\linewidth]{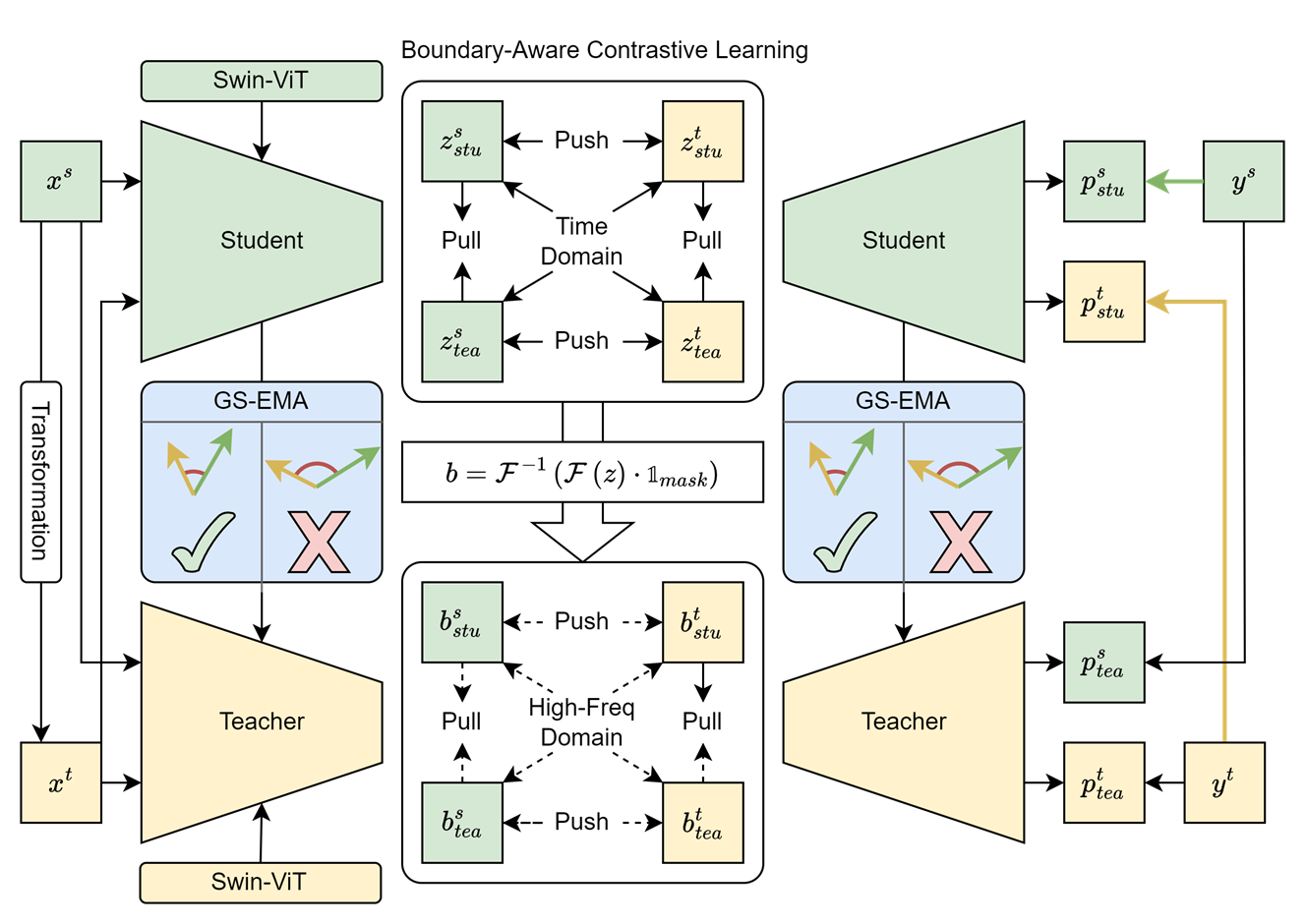}}
\caption{Schematic of the proposed model.}
\label{fig2}
\end{figure}

\vspace{-0.5cm}
Fig. \ref{fig2} depicts our neural network architecture dedicated for domain generalization tasks. It initiates with 3D source images, which undergoes image transformation to produce target images. Once both the source and target images are obtained, they are separately fed into the encoders of the student and teacher networks.

After acquiring the latent space features, a boundary-aware contrastive learning loss is computed. The central notion here is to amalgamate the same instance subjected to diverse transformations, while distancing different instances, aiming to grasp instance-aware representations. This contrastive learning differs from transformation predictions, as it strives to attain transformation-invariant representations. The latent space features are then decoded to yield predictions, which are supervised using ground truth. 

Within the student network, the green arrow signifies fully supervised learning for the source images, and the yellow arrow represents the same for the target images. By analyzing the gradient relationship between these two losses, a novel GS-EMA strategy is devised to update the parameters of teacher network. If the gradient angle between the losses is less than 90 degrees, it indicates that the network has learned domain-invariant features, prompting an EMA update. Conversely, if the gradient angle exceeds 90 degrees, no EMA update is performed, as this suggests the network has grasped domain-specific features, which is not conducive to domain generalization tasks. Ultimately, after several updates, a teacher network enriched with more domain-invariant features is achieved, readying it for domain generalization tasks.

\subsection{Problem Definition and Data Transformation}

Let $\mathcal{X}$ be the input (image) space and $\mathcal{Y}$ be the
segmentation (label) space, a domain is defined as a joint distribution $P_{XY}$ on $\mathcal{X} \times \mathcal{Y}$.
In the context of DG, we have access to $K$ similar but distinct source domains 
$\left \{ \left ( x^{k}_s,y^{k}_s \right )  \right \}  _{k=1}^{K}$, each associated with a joint distribution $P_{XY}^{k}$. Note that $P_{XY}^{i} \neq P_{XY}^{j}$ with $i \neq j$ and $i, j \in \{1, ... , K\}$.  The goal of DG is to learn
a predictive model using only source domain data such that the prediction error on an unseen target domain is minimized.

To enhance the model adaptability to previously unseen data domains, we use data transformations to simulate the distribution of the target domain data. The simulated target data is represented as $\left \{ \left ( x_{k}^{t},y_{k}^{t} \right ) \right \}  _{k=1}^{K}$.
The process of data transformation encompasses several key steps, including geometric transformations, intensity alterations, noise injection and smoothing, histogram shifting, as well as bias field correction. These operations collectively aim to generate diverse target data, empowering the model with enhanced generalization capabilities to adapt to various data sources and target domains.

\vspace{-0.3cm}
\subsection{Gradient Surgery Exponential Moving Average}

In a teacher-student network setup, when the student network is tasked with learning from data originating from different domains, we calculate distinct losses for each domain in Eq.~\ref{Lsrc}. This allows us to obtain gradient information specific to each domain. Our fundamental hypothesis is that when the angle between gradients from different domains is less than 90 degrees, it suggests that the student network has effectively learned how to extract domain-invariant features. In such cases, we employ EMA to update the parameters of student network, subsequently transferring these parameters to the teacher network. This transfer is performed to better capture universal features.

However, when the angle between gradients from different domains exceeds 90 degrees, it indicates that the student network is primarily focused on learning domain-specific features. In such scenarios, we abstain from utilizing EMA for parameter updates and refrain from transmitting these parameters to the teacher network. This strategic approach ensures that the student network can efficiently discriminate between features originating from different domains, enabling it to adapt effectively to the challenges of multi-task learning.
\begin{equation}
\label{Ldce}
\begin{split}
\mathcal{L}_{DCE} (p, y)
&= \frac{1}{N} \sum_{i=1}^{N} \left ( 1 -  \frac{2\left | p \cap y \right | }{\left |  p \right | + \left | y \right |} -  y\log p\right)  
\end{split}
\end{equation}
\begin{equation}
\label{Lsrc}
\begin{split}
\mathcal{L}_{stu}^{src} = \mathcal{L}_{DCE} (p_{stu}^{s}, y^{s}) , \  \mathcal{L}_{stu}^{trg} = \mathcal{L}_{DCE} (p_{stu}^{t}, y^{t}) 
\end{split}
\end{equation}

% \begin{algorithm}
% \caption{Gradient Surgery Exponential Moving Average}
% \KwData{Student network parameters $\theta_{stu}$; 
% Teacher network parameters $\theta_{tea}$;
% Loss on source data in student network $\mathcal{L}_{src}$;
% Loss on target data in student network $\mathcal{L}_{trg}$;
% EMA decay coefficient $\alpha$.
% }
% \KwResult{Decide whether updated $\theta_{tea}$ with EMA from $\theta_{stu}$. }

% \For{each mini-batch}{

% $\nabla  \mathcal{L}_{stu}^{src} \rightarrow g_{src} $\;
    
% $\nabla  \mathcal{L}_{stu}^{trg} \rightarrow g_{trg}  $\;

%     \uIf{$\langle g_{src},g_{trg}\rangle \leq 0$}{

%     \[\theta_{tea}' = \theta_{tea} \cdot \alpha + (1 - \alpha) \cdot \theta_{stu} \]
        
%     }\ElseIf{$\langle g_{src},g_{trg}\rangle > 0$}{
%     \[\theta_{tea}' = \theta_{tea} \]

%     }
    
% Update $\theta_{tea}$ $\theta_{stu}$ 

% }
% \label{algo}
% \end{algorithm}

\begin{algorithm}[]
\caption{Gradient Surgery Exponential Moving Average}
\KwData{Student network parameters $\theta_{stu}$; 
Teacher network parameters $\theta_{tea}$;
Loss on source data in student network $\mathcal{L}_{src}$;
Loss on target data in student network $\mathcal{L}_{trg}$;
EMA decay coefficient $\alpha$.}
\KwResult{Decide whether updated $\theta_{tea}$ with EMA from $\theta_{stu}$. }
\For{each mini-batch}{
    $\nabla  \mathcal{L}_{stu}^{src} \rightarrow g_{src}$\;
    $\nabla  \mathcal{L}_{stu}^{trg} \rightarrow g_{trg}$\;
    \uIf{$\langle g_{src},g_{trg}\rangle \leq 0$}{
        $\theta_{tea}' = \theta_{tea} \cdot \alpha + (1 - \alpha) \cdot \theta_{stu}$\;
    }\ElseIf{$\langle g_{src},g_{trg}\rangle > 0$}{
        $\theta_{tea}' = \theta_{tea}$\;
    }
    Update $\theta_{tea}$ with $\theta_{tea}'$\;
    Update $\theta_{stu}$ as needed\;
}
\end{algorithm}

\vspace{-0.5cm}
\subsection{Boundary-Aware Contrastive Learning}
In our study, we tackle the challenge of uneven distribution of classes in the segmentation of aneurysms by proposing a unique contrastive learning approach that operates within a teacher-student network configuration. This method enhances the distinction between matching (positive) and non-matching (negative) sample pairs by employing a Fourier transformation strategy, which is particularly adept at isolating high-frequency elements that delineate boundaries. Transitioning from volume-based to boundary-based analysis ensures that the presence of small aneurysms is not disproportionately low compared to larger vessels.

Both the student and teacher branches receive two distinct sets of data: the original data from the source domain, represented as \(x^{s}\), and the corresponding transformed data \(x^{t}\). Consequently, the latent feature representations from the student network are symbolized as \(z_{stu}^{s}\) and \(z_{stu}^{t}\), while those from the teacher network are signified as \(z_{tea}^{s}\) and \(z_{tea}^{t}\).

Advancing further, we harness the power of Fourier transformation paired with a high-frequency filter in Eq. \ref{fourier} to extract features that are cognizant of the boundaries within the data. These extracted features from both student and teacher networks are represented as \(b_{stu}^{s}\), \(b_{stu}^{t}\), \(b_{tea}^{s}\), and \(b_{tea}^{t}\) respectively. Our primary objective within this feature space is to cultivate instance-specific representations that are closely aligned when the same instance is encoded differently, while simultaneously ensuring a clear demarcation between distinct instances, irrespective of the encoder used.

To clarify the relationships within our contrastive learning framework, we delineate the instances processed through different encoders as positive pairs when they originate from the same instance. This includes pairs like \(z_{stu}^{s}\) with \(z_{tea}^{s}\), and \(z_{stu}^{t}\) with \(z_{tea}^{t}\). In contrast, negative pairs consist of different instances that have been encoded either by the same or by different encoders, such as \(z_{stu}^{s}\) with \(z_{stu}^{t}\), and \(z_{tea}^{s}\) with \(z_{tea}^{t}\), as well as cross-encoder pairs like \(z_{stu}^{s}\) with \(z_{tea}^{t}\), and \(z_{tea}^{s}\) with \(z_{stu}^{t}\). These delineations form the basis of our contrastive learning process.

Moving forward, we apply a Fourier transformation to the volume features to construct an amplitude map, which is crucial for identifying the salient high-frequency components that highlight boundaries in Eq.\ref{fourier}. A specialized square mask is then utilized to isolate this high-frequency information in Eq.\ref{fourier}. Here, the Fourier transform and its inverse are denoted by \(\mathcal{F}\) and \(\mathcal{F}^{-1}\) respectively. The mask $\mathbb{1}_{ mask }$, with value zero in its center and one at the periphery, has the same shape as $z$.

For boundary features, positive pairs are formed by analogous instances across the student and teacher networks, such as \(b_{stu}^{s}\) with \(b_{tea}^{s}\), and \(b_{stu}^{t}\) with \(b_{tea}^{s\rightarrow t}\). Conversely, negative pairs are created by combining features from distinct instances, which may be within the same network or across both, exemplified by pairs such as \(b_{stu}^{s}\) with \(b_{stu}^{t}\), and \(b_{tea}^{s}\) with \(b_{tea}^{t}\), as well as inter-network pairs like \(b_{stu}^{s}\) with \(b_{tea}^{t}\), and \(b_{tea}^{s}\) with \(b_{stu}^{t}\).
\begin{equation}
\label{fourier}
b = \mathcal{F}^{-1}  \left( \mathcal{F} \left( z \right) \cdot \mathbb{1}_{mask}  \right)
\end{equation}
\begin{equation}
\label{huv}
h(u, v)=\frac{u^{T} v}{\|u\|_{2}\|v\|_{2}}
\end{equation}
\begin{equation}
\label{ltranswarp}
\mathcal{L}_{c}=-\log \frac{\sum_{i=1}^{N_p}e^{{\left(h\left(u_{i}^+, v_{i}^{+}\right)  \right)}}}{\sum_{i=1}^{N_p}e^{{\left(h\left(u_{i}^+, v_{i}^{+}\right)  \right)}} + \sum_{j=1}^{N_n}e^{{\left(h\left(u_{j}^+, v_{j}^{-}\right)  \right)}}}
\end{equation}

To quantify the similarity of these pairs, we compute the cosine similarity for each within both the time and frequency domains in Eq.\ref{huv} and Eq.\ref{ltranswarp}. The similarity for positive pairs is expressed as \(h(u_{i}^+, v_{i}^+)\) where \(i\) spans all positive pair indices, and the similarity for negative pairs is articulated as \(h(u_{j}^+, v_{j}^-)\) where \(j\) represents the indices of all negative pairs. Here, \(N_p\) stands for the count of positive pairings, and \(N_n\) corresponds to the count of negative pairings.

The contrastive learning loss for these high-frequency boundary pairs is then calculated using the same equation as for the volumetric pairs. By summing up the volumetric contrastive learning loss with the boundary contrastive learning loss, we derive a comprehensive boundary-aware contrastive learning loss. This loss function is designed to finely tune our model to discriminate between the nuanced features of aneurysms, enhancing its segmentation performance.

\vspace{-0.3cm}
\subsection{Overall framework and training objective}
The loss function consists of two parts. $\mathcal{L}_{DCE}$ fully supervises the four outputs of the teacher and student networks. BACL includes volume contrast $\mathcal{L}_{c}^{z}$ and boundary contrast $\mathcal{L}_{c}^{b}$. The ratio of $\lambda_1$ to $\lambda_2$ is set at 0.25:0.5.
\begin{equation}
\mathcal{L}=\lambda_1 \cdot (\mathcal{L}_{stu}^{src}+\mathcal{L}_{stu}^{trg}+\mathcal{L}_{tea}^{src} +\mathcal{L}_{tea}^{trg}) + \lambda_2 \cdot (\mathcal{L}_{c}^{z} + \mathcal{L}_{c}^{b})
\end{equation}

\section{Experiments}

\subsection{Experimental Setting}
\textbf{Dataset}:
We tested our method with 3DRA images from 223 patients from the @neurIST dataset \cite{benkner2010neurist}. These images were collected from four distinct medical institutions, each employing varied scanning equipment and imaging protocols. Consequently, this dataset exhibits a broad various in both visual characteristics and resolution. The data diversity can evaluate the robustness and adaptability of our proposed GS-EMA method.

\textbf{Implementation details}:
Our study was conducted on a NVIDIA RTX 3090 GPU. We utilized the Swin-UNet \cite{cao2022swin} architecture for both the student and teacher networks in our framework. The training was set to 100 epochs. To determine whether to apply EMA updates, we experimented with setting the EMA coefficient \(\alpha\) to either 0.9999 or 0.9. We started with an initial learning rate of 0.001 and adjusted it downwards by multiplying by 0.1 after every ten epochs.
The code will be publicly available soon.

\vspace{-0.3cm}
\subsection{Quantitative Results}
\vspace{-0.3cm}
\begin{table}[htb!]
    \centering
    \resizebox{\linewidth}{!}{
    \begin{tabular}{c|c|c|c|c} \hline \hline 
         &  DSC (\%) $\uparrow$ &  Sen (\%) $\uparrow$& Jac (\%) $\uparrow$&VS (\%) $\uparrow$\\ \hline 
         nnUNet \cite{isensee2021nnu}& 59.61 & 57.51 & 47.38 &70.91\\ 
         VASeg \cite{lin2023high}& 60.28 &54.47 &49.82 &67.91\\ 
         FedDG \cite{liu2021feddg}& 64.50 &64.31 &54.26 &74.73\\ 
         CMDG \cite{ouyang2022causality}& 65.01  & 64.10 & 54.11 &73.38\\ 
         Ours& \textbf{71.89} &\textbf{70.88}  & \textbf{62.36} &\textbf{80.00}\\  \hline \hline 
         no EMA&  61.52&  55.64&  50.91&69.01\\ 
         EMA&  64.71&  62.86&  54.03&72.64\\ 
         GS-EMA& \textbf{68.49}& \textbf{72.79}& \textbf{58.40}&\textbf{76.63}\\  \hline 
         BACL-V & {68.49}& {72.79}& {58.40}&{76.63}\\ 
         BACL-B& 70.62& \textbf{75.14}& 60.54&78.22\\
         BACL& \textbf{71.89} &70.88  & \textbf{62.36} &\textbf{80.00}\\ 
         \hline \hline 
    \end{tabular}}
    \caption{Quantitative results including compare with SOTAs and ablation studies. Critical metrics includes the Dice similarity coefficient (DSC), Sensitivity (Sen), Jaccard index (Jac) and Volume similarity (VS).}
    \label{tab:1}
\end{table}

Table.\ref{tab:1} includes comparison with state-of-the-art (SOTA) methods and two ablation studies.
Our model outperforms traditional segmentation approaches like nnUNet \cite{isensee2021nnu} and aneurysm-focused VASeg \cite{lin2023high}, as well as domain-generalizing methods for medical image segmentation, including CMDG \cite{ouyang2022causality} and FedDG \cite{liu2021feddg}.
The ablation study highlights that our GS-EMA algorithm, regulating EMA updates with gradient relation, surpasses both regular and non-EMA methods in segmenting aneurysms. It also indicates superior results for BACL when integrating volume (BACL-V) and boundary (BACL-B) learning, compared to using either alone.

\vspace{-0.3cm}
\subsection{Visual Inspection}
\begin{figure}[!htb]
\centerline{\includegraphics[width=\linewidth]{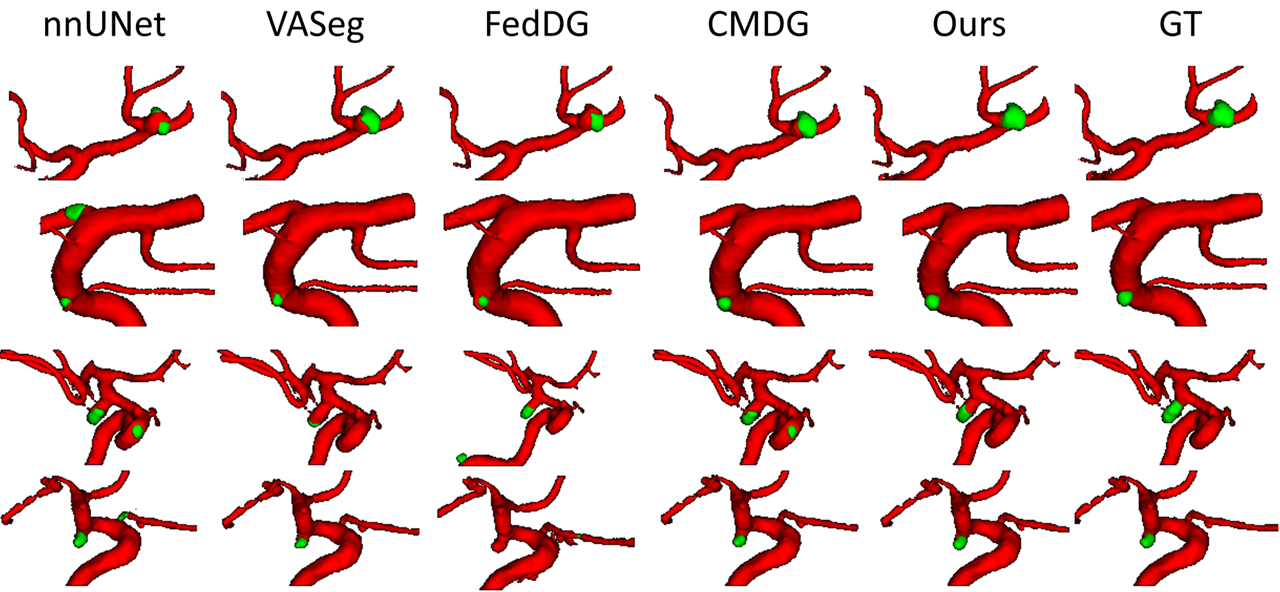}}
\caption{Comparative visualization of SOTAs and ours method on aneurysm segmentation.}
\label{fig3}
\end{figure}

Fig.~\ref{fig3} offers a visual comparison of aneurysm segmentation between our method and the SOTAs. It is evident from the comparison that our approach is less prone to over-segmentation while also being able to segment aneurysms more completely.

\begin{figure}[!ht]
\centerline{\includegraphics[width=\linewidth]{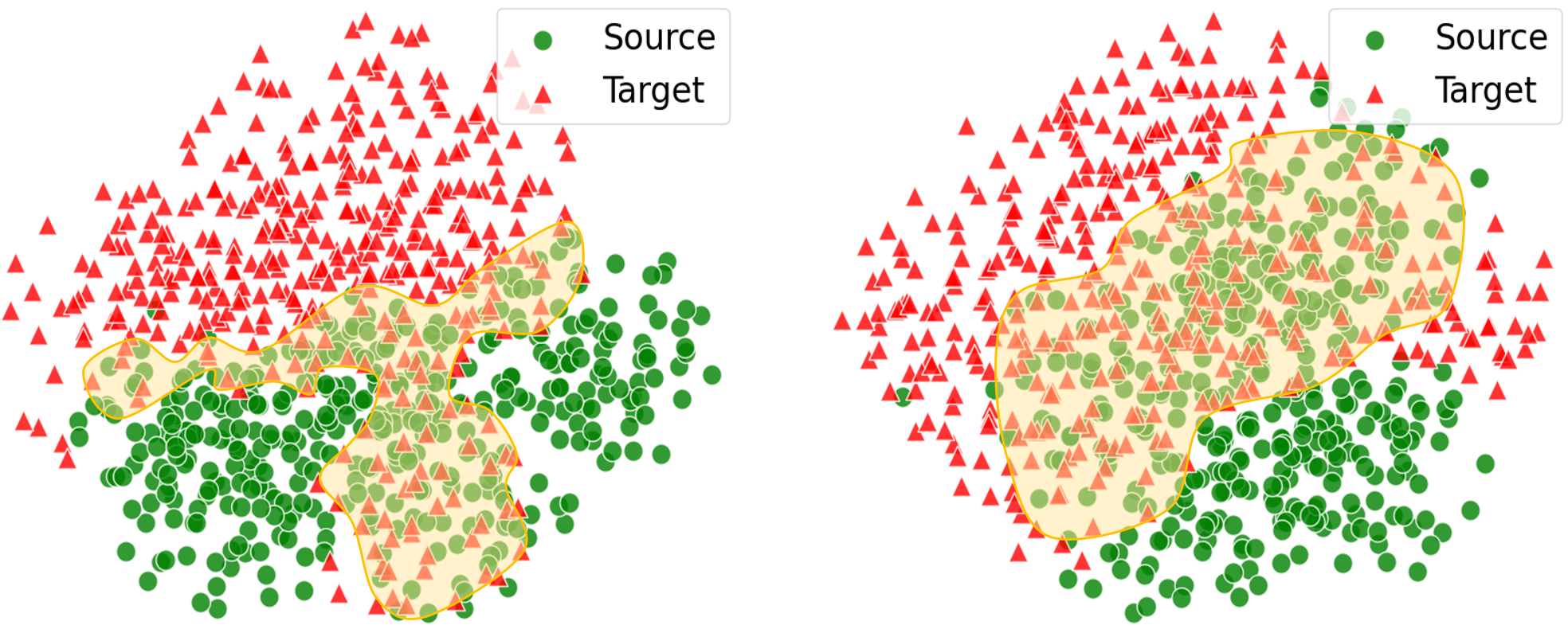}}
\caption{The t-SNE visualization of latent features from EMA (left) and GS-EMA (right).}
\label{fig4}
\end{figure}

\vspace{-0.3cm}
Fig.~\ref{fig4} 
displays a t-SNE comparison of latent features using EMA and GS-EMA. The larger overlap achieved by GS-EMA indicates a stronger capability of the model to extract domain-invariant features.

\vspace{-0.5cm}
\section{Conclusion}
In summary, our study introduces an effective GS-EMA algorithm and a boundary-aware contrastive learning technique for aneurysm segmentation. These methods outperform existing approaches by minimizing over-segmentations and capturing more complete aneurysm structures. 
For future work, we plan to apply our GS-EMA technique to a wider array of medical imaging datasets for further validation and enhancement.

\bibliographystyle{IEEEbib}
\bibliography{refs}

\end{document}